\documentclass[conference,10pt,letterpaper,twoside]{style/ieeeconf}

\IEEEoverridecommandlockouts    %
\makeatletter 

\let\proof\@undefined
\let\endproof\@undefined

\let\NAT@parse\undefined
\makeatother

\usepackage[x11names]{xcolor}

\usepackage[ruled, vlined, linesnumbered]{algorithm2e}
\makeatletter
\newcommand{\removelatexerror}{\let\@latex@error\@gobble}
\makeatother
\SetNlSty{bfseries}{\color{black}}{}
\usepackage{xcolor}

\SetCommentSty{mycommfont}

\usepackage[numbers, sort, compress]{natbib}
\usepackage{graphicx}
\usepackage{amsmath,amssymb}
\usepackage{caption}
\usepackage{subcaption}
\usepackage[export]{adjustbox}

\usepackage{amsthm}  

\usepackage{style/perl_acronyms}
\usepackage{style/perl_math}
\usepackage{style/perl_SIunits}
\usepackage{style/perl_misc}
\usepackage{multirow}
\usepackage{verbatimbox}
\usepackage[normalem]{ulem}
\usepackage{makecell}
\usepackage{array}
\newcolumntype{C}[1]{>{\centering\arraybackslash}p{#1}}

\usepackage{hyperref}

\usepackage{flushend}

\usepackage[normalem]{ulem}

\begin{document}

\title{A Preview of HoloOcean 2.0}
    \author{{Blake Romrell, Abigail Austin, Braden Meyers, Ryan Anderson, Carter Noh, and Joshua G. Mangelson}
      \thanks{This work was funded under Department of Navy awards N00014-24-1-2301 and N00014-24-1-2503 issued by the Office of Naval Research. Parts of this work were also funded by Naval Sea Systems Command (NAVSEA), Naval Surface Warfare Center - Panama City Division (NSWC-PCD) and Naval Undersea Warfare Center - Keyport Division (NUWC-KPT) under the Naval Engineering Education Consortium (NEEC) Grant Program under award numbers N00174-23-1-0005 and N00178-23-1-0006.}%
      \thanks{B. Romrell, A. Austin, B. Meyers, C. Noh, and J. Mangelson are all at Brigham Young University. They can be contacted at: \texttt{\{romrellb, abiausti, bjm255, cartern2, mangelson\}@byu.edu}~. Ryan Anderson is at Lucid Software \texttt{ryananderson@lucidchart.com}}
    }
\maketitle

\begin{abstract}
Marine robotics simulators play a fundamental role in the development of marine robotic systems. With increased focus on the marine robotics field in recent years, there has been significant interest in developing higher fidelity simulation of marine sensors, physics, and visual rendering capabilities to support autonomous marine robot development and validation. 
HoloOcean 2.0, the next major release of HoloOcean, brings state-of-the-art features under a general marine simulator capable of supporting a variety of tasks.
New features in HoloOcean 2.0 include migration to Unreal Engine (UE) 5.3, advanced vehicle dynamics using models from Fossen, and support for ROS2 using a custom bridge.
Additional features are currently in development, including significantly more efficient ray tracing-based sidescan, forward-looking, and bathymetric sonar implementations; semantic sensors; environment generation tools; volumetric environmental effects; and realistic waves.
\end{abstract}

\acresetall
\IEEEpeerreviewmaketitle

\section{Introduction}
\label{sec:intro}

Marine robotics simulators have supported research and development for autonomous underwater and surface vessels for several decades.
Simulation is a critical capability that enables development of algorithms for navigation, perception, manipulation, and control, as well as validation of real-world systems and missions. 

As the field of underwater robotics has grown, so has the need for high-fidelity simulations.
An increased emphasis on vision-based algorithms in mobile robotics has pushed simulators toward photorealistic graphics renderings. 
Complex missions involving multiple agents or long duration operations require well-modeled dynamics to ensure accurate results. 

In the last several years, marine robotics simulators have seen significant advancements to meet these needs.
At least six new simulators have been released in the last six years \cite{LRAUV, MARUS, UNav-Sim, Suzuki, SMaRC, DAVE, stonefish}.
Some simulators such as LRAUV \cite{LRAUV} and Stonefish \cite{stonefish} focus on precise vehicle dynamics.
Others such as MARUS \cite{MARUS} and UNavSim \cite{UNav-Sim} leverage the high-quality visual rendering available from modern game engines such as Unity \cite{unity} and Unreal Engine \cite{unrealengine} to enable vision-based algorithms and artificial intelligence. 

The HoloOcean simulator was released in 2022, with the objective of providing high-fidelity visuals and sensor models to enable algorithm development for marine robot navigation, perception, estimation, and localization \cite{potokarHoloOcean2022, potokarHoloOceanSonar2022}.
It emphasized detailed simulation of sonar sensors, including sidescan, imaging, and bathymetric sonars. 
In the years since its release, HoloOcean has been utilized by researchers at universities and government agencies across the world.  

In this paper, we give a preview of HoloOcean 2.0, a major update to HoloOcean.
HoloOcean 2.0 brings more state-of-the-art features into a single, high-fidelity and user friendly simulator and introduces novel features not found in other simulators.  

\begin{figure}[t]
    \centering
    \includegraphics[width=\linewidth]{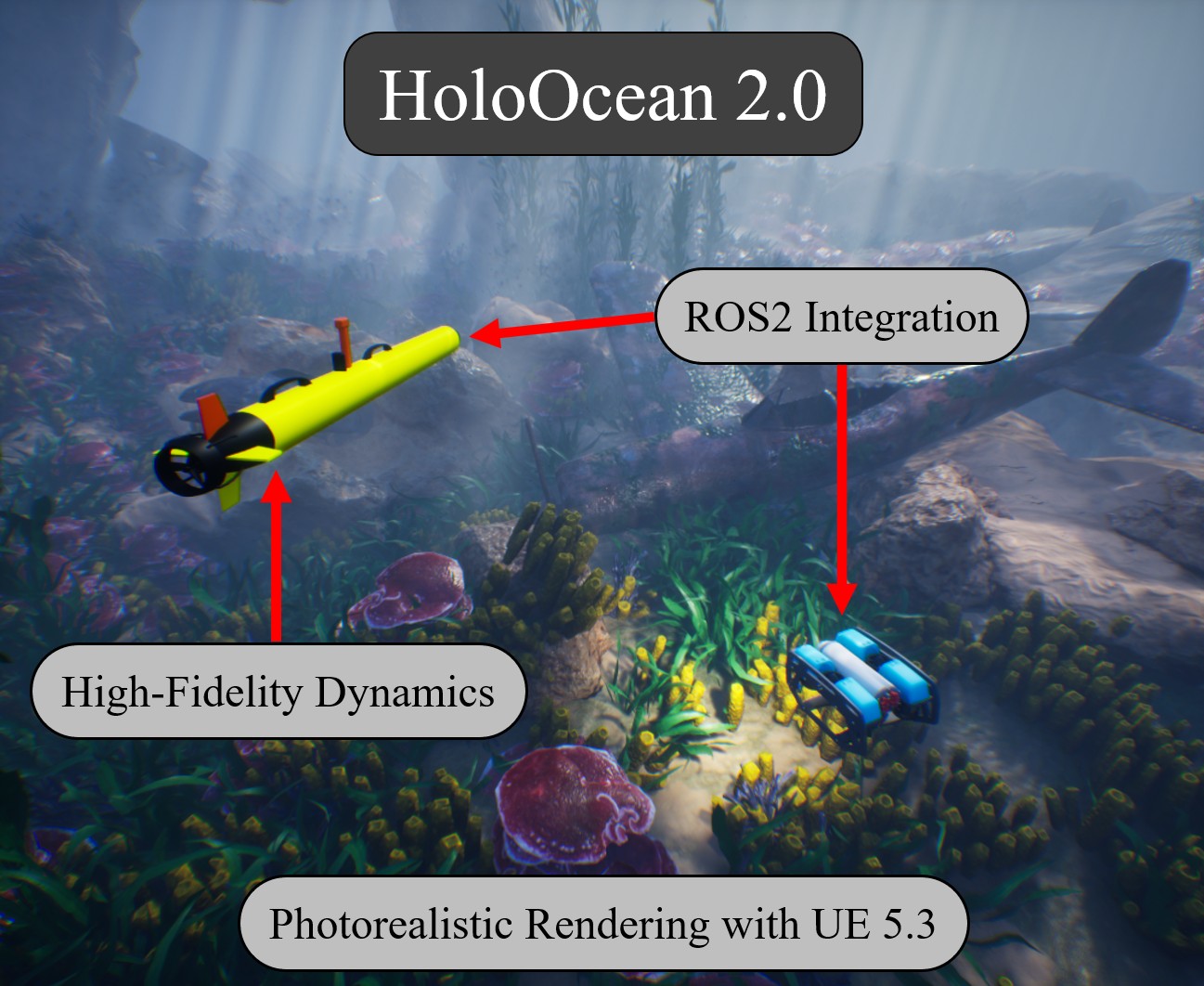}
    \caption{HoloOcean 2.0 incorporates state-of-the-art features such as enhanced visuals with UE 5.3, support for ROS2, and advanced vehicle dynamics. }
    \label{fig:main}
    \vspace{-1em}
\end{figure}

The paper is organized as follows. Section \ref{sec:new-features} describes the following new features available in HoloOcean 2.0, including:
\begin{itemize}
    \item migration to Unreal Engine (UE) 5.3,
    \item improved vehicle dynamics,
    \item support for ROS2, and 
    \item integration of the BlueROV and CoUG-UV vehicles.
\end{itemize}

Section \ref{sec:in-development} provides details on features and improvements currently in development, including:
\begin{itemize}
    \item an improved sonar implementation using ray casting,
    \item semantic labeling for camera and sonar sensors,
    \item automatic environment generation,
    \item volumetric environment effects, and 
    \item accurate wave simulation for visuals and dynamics.
\end{itemize}

Section \ref{sec:conclusion} concludes the paper and discusses HoloOcean's place in the future of underwater robotics simulation.

\begin{figure*}
    \centering
    \includegraphics[width=\linewidth]{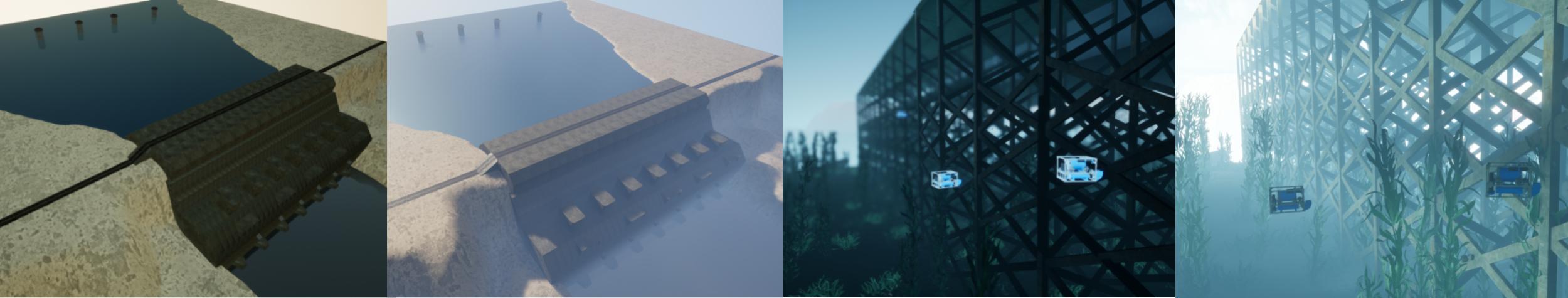}
    \caption{Comparison of environment rendering in UE 4.27 and UE 5.3. From left to right: (1) original Dam environment in UE 4.27, (2) upgraded Dam environment in UE 5.3, (3) Pier Harbor environment in UE 4.27, (4) Pier Harbor environment in UE 5.3.}
    \label{fig:environment-improvement}
    \vspace{-2em}
\end{figure*}

\section{New Features in HoloOcean 2.0}
\label{sec:new-features}

New features available in HoloOcean 2.0 include a migration to Unreal Engine 5.3 to take advantage of its enhanced visual rendering; high-fidelity vehicle dynamics for the torpedo vehicles; support for ROS2; and new vehicles. 

\subsection{Migration to Unreal Engine 5.3}
HoloOcean has been upgraded from UE version 4.27 to version 5.3 to access new features added in UE 5.
Unreal has added the Lumen lighting system \cite{Lumen} for dynamic global lighting and reflections, which has greatly improved the realism of HoloOcean environments.
In addition, HoloOcean can now support much larger worlds due to the new Nanite Virtualized Geometry \cite{Nanite} system utilized by UE 5. 
Nanite renders much larger and more detailed scenes than those possible in UE 4, is easily applied to meshes, and automatically generates Level of Detail for rendering.
By leveraging Lumen and Nanite, HoloOcean can now simulate larger, more detailed, and more realistic environments. Figure \ref{fig:environment-improvement} shows these improved visuals for two HoloOcean environments.

\subsection{Improved Vehicle Dynamics}
Precise vehicle dynamics are increasingly important in marine robotics as users seek to close the sim-to-real gap. 
Similarly to other recent simulators such as LRAUV \cite{LRAUV} and UNav-Sim \cite{UNav-Sim}, HoloOcean 2.0 addresses this need by implementing Thor Fossen's high-fidelity dynamics models for torpedo-style vehicles \cite{fossen}. 
These models represent the current state-of-the-art for marine vehicle simulation, with accurate modeling of hydrostatic forces, dissipative forces, system inertia, and control surface dynamics and effects.
HoloOcean uses Fossen's models to generate accelerations for each vehicle based on the vehicle's state and input at each tick. These accelerations are then passed into UE 5's physics engine to handle collisions and other external forces. 

HoloOcean 2.0's implementation of Fossen dynamics consists of 1) a vehicle controller written in Python that implements customized dynamics for each agent, and 2) a dynamics manager that interfaces each agent's vehicle controller with the rest of the HoloOcean simulation.
The vehicle controller for each agent can be configured with custom parameters during scenario configuration to match a specific vehicle and is equipped with built-in depth and heading control.
The default model parameters are for the REMUS 100 \cite{HII_REMUS100}. A summary of the configurable parameters for each vehicle is given in Table \ref{tab:fossen-params}. 

Currently, only Fossen's models for torpedo-style vehicles are implemented, but Fossen has also defined models for surface and hovering-style vehicles. 
Implementations of these in HoloOcean are currently in development. 
Users can also create vehicle controllers for their own custom dynamics that can be used with the dynamics manager. 

 \begin{table}[t]
    \centering
    \renewcommand{\arraystretch}{1.3}
    \begin{tabular}{| >{\centering\arraybackslash}m{2cm} | >{\centering\arraybackslash}m{5.5cm} |}
        \hline
        \textbf{Category} & \textbf{Parameters} \\
        \hline
        Environmental & Water Density, Gravity, Currents \\
        \hline
        Physical & Mass, Length, Diameter, Inertia \\
        \hline
        Hydrodynamic  & Low-speed Linear Damping \\
        \hline
        Hydrostatic  & Center of Bouyancy and Mass locations \\
        \hline
        Control Surfaces & Time Constants, Lift/Thrust Coefficients, Fin Positions, Fin Area \\
        \hline
        Autopilot & Pitch-Depth PID Gains, Heading SMC Gains \\
        \hline
    \end{tabular}
    \caption{List of configurable parameters for HoloOcean's Fossen vehicle dynamics.}
    \label{tab:fossen-params}
    \vspace{-2em}
\end{table}

\subsection{ROS2 Bridge}
\begin{figure}[b]
    \centering
    \includegraphics[width=\linewidth]{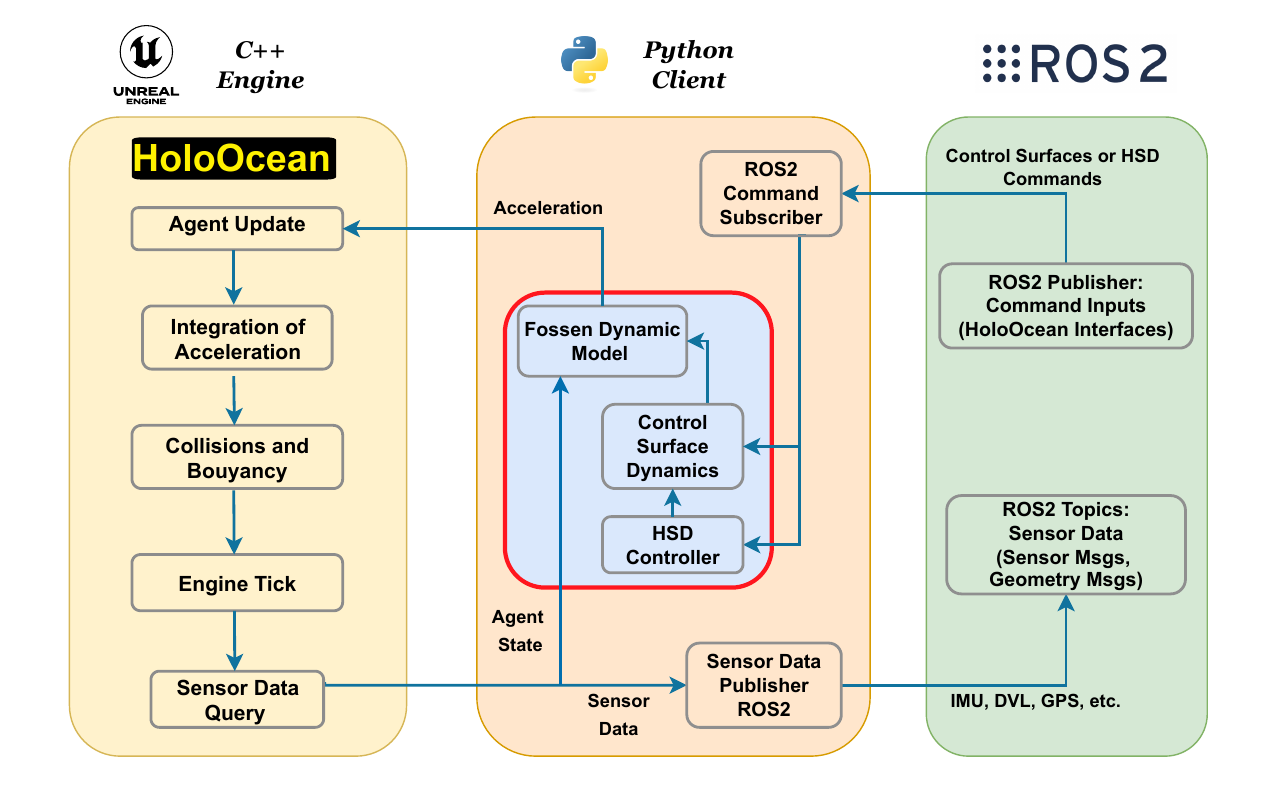}
    \caption{Flow diagram demonstrating the the use of Fossen dynamic and the HoloOcean ROS2 bridge in HoloOcean 2.0. }
    \label{fig:fossen}
\end{figure}

HoloOcean's initial release supported external system integration through LCM, which was selected for its light weight.
HoloOcean 2.0 builds on this by adding support for ROS2, which offers more features and a larger active user base.
HoloOcean users can now use a custom ROS2 bridge to publish sensor data and send commands from HoloOcean to an external network. 
Several example ROS2 Python nodes are included in HoloOcean's documentation to demonstrate how a user can create their own ROS2 interface for their specific scenario.

Figure \ref{fig:fossen} shows the flow of information between Unreal Engine, HoloOcean, and ROS. 
During simulation, sensor data from the agent is received by HoloOcean's Python client and translated into a ROS2 message format (sensors messages, geometry messages, and custom HoloOcean messages), which can then interact with other agents external to HoloOcean.

The ROS2 bridge has been configured specifically to work with the implementation of Fossen dynamics as described above. Vehicle commands (actuator positions, controller setpoints for speed, depth, and heading, etc.) can be sent to a Fossen-controlled torpedo agent through the Fossen dynamics manager.

\begin{figure}[b]
    \centering
    \includegraphics[width=\linewidth]{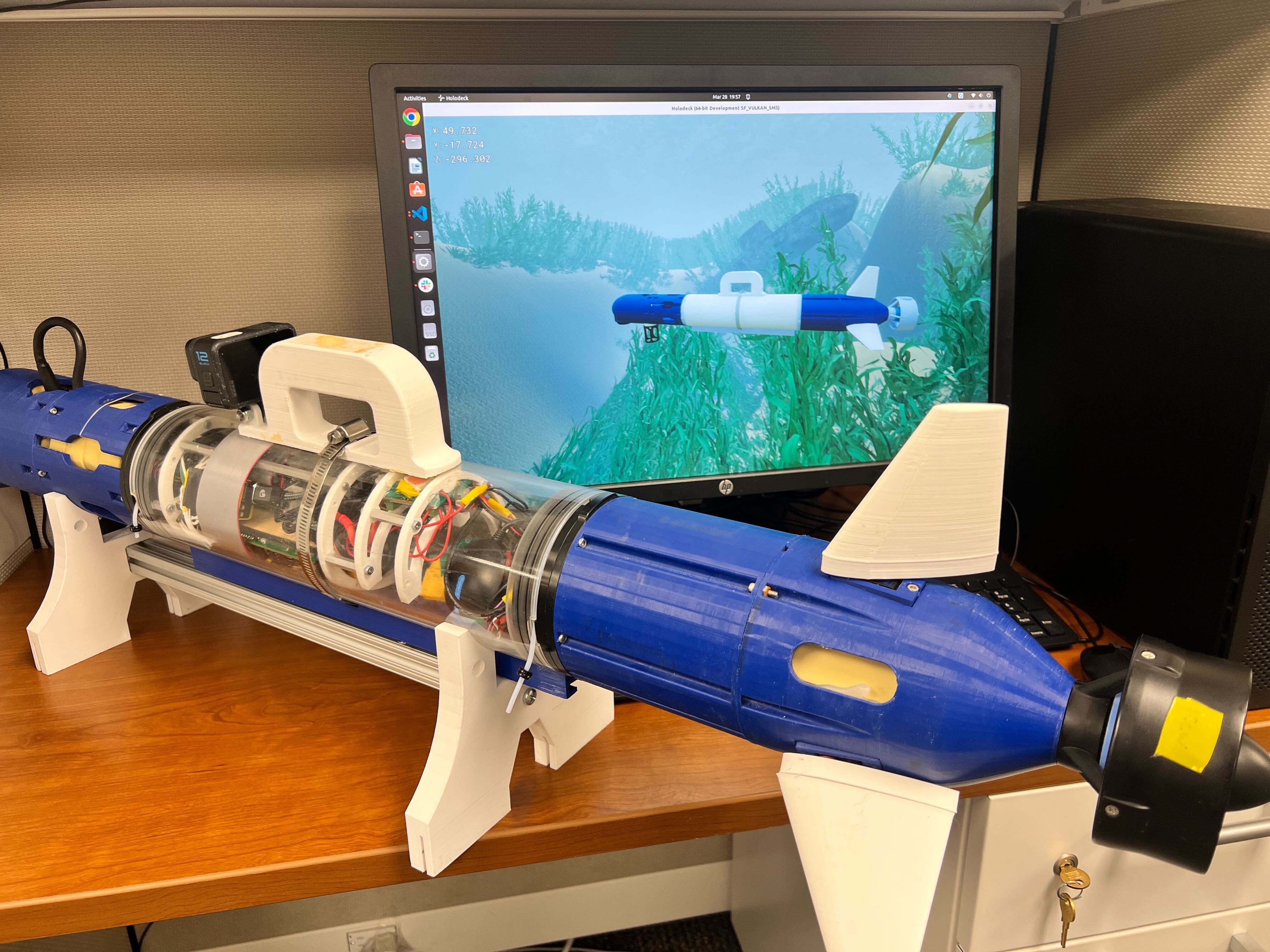}
    \caption{The HoloOcean ROS2 bridge can be used to perform hardware-in-the-loop simulations as demonstrated above.}
    \label{fig:hardware-in-loop}
\end{figure}

The BYU FRoST Lab uses the ROS2 bridge to run hardware-in-the-loop simulations to verify the functionality of control, navigation, and localization algorithms with the simulated ROS2 sensor drivers. This is shown in Figure \ref{fig:hardware-in-loop}.

\subsection{New Vehicles}
\begin{figure}[t]
    \centering
    \includegraphics[width=\linewidth]{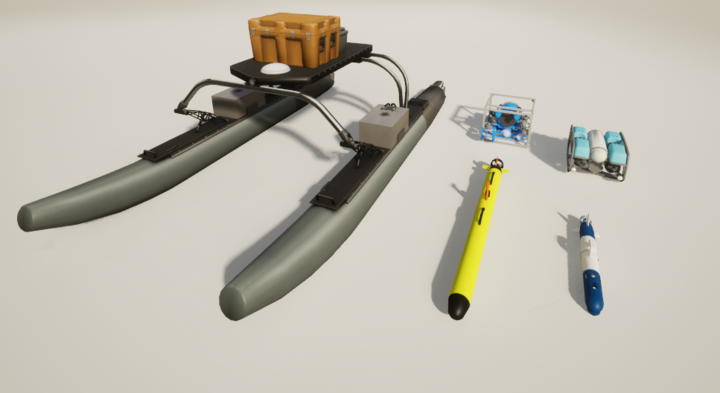}
    \caption{Marine robotics platforms available in HoloOcean. From left to right: a surface vessel, Hovering AUV, Torpedo AUV, BlueROV2, and CoUG-UV.}
    \label{fig:vehicles}
    \vspace{-2em}
\end{figure}

HoloOcean's first launch featured three marine vehicles: a custom hovering AUV vehicle, a torpedo vehicle based on the L3Harris IVER3 \cite{IVER3}, and a surface vessel based on the WAM-V \cite{WAM-V}.
HoloOcean 2.0 introduces two new ready-to-use vehicles.
First, we have added the BlueROV2 Heavy from Blue Robotics \cite{BlueROV}, a widely used hovering vehicle that can be equipped with a variety of sensors.
Second, we include the CoUG-UV, a custom small-scale torpedo vehicle developed by the BYU FRoST Lab for large-scale multi-agent experiments.
Figure \ref{fig:vehicles} shows all of HoloOcean's available marine platforms. Additional ground and aerial vehicles (such as a quadrotor) are also available. 

\section{Features in Development}
\label{sec:in-development}

Several additional features are currently in development for future releases. These will be detailed in the following subsections. 

\subsection{New Sensors}
We are working on adding several new sensor models to HoloOcean. These sensors are fully implemented and are currently awaiting final testing and documentation. 

First, we have added a higher-fidelity camera model. HoloOcean's original RGB camera sensor only had parameters to adjust the image size. Our new camera sensor adds 29 new parameters, including field of view angle, shutter speed, focal distance, and more. These allow the user to configure their camera to match a specific real-world sensor. 
Second, we have implemented a depth camera. The depth camera has the same configurable parameters as the new camera implementation, but also returns a depth channel.
Finally, we have implemented a LiDAR sensor, adapted from the Carla simulator \cite{Dosovitskiy17}. The LiDAR has parameters for number of lasers, rotation frequency, field of view angles, maximum range, and more.

\subsection{Ray Casting Sonar}
HoloOcean's original sonar simulation method relies on a cached representation of the environment via octrees. At runtime, the sonar sensor queries these octrees to recover the value that the sonar ray would ``see" based on acoustic ray theory and the projection model for the specific sonar sensor type.
It was expected that ray tracing directly would be too computationally expensive to use in real-time, but by caching surfaces in the octree structure, we are able to trade higher memory usage for less computational load at runtime, improving simulation speed. 

However, the octree implementation has several downsides. 
The initial simulation that generates the octrees is extremely computationally intensive and octrees must be regenerated if the environment changes or certain sonar configuration parameters are updated. 
These problems can significantly hinder development and usage of new environments for sonar applications. 
After initial octree generation, memory-access time to load the cached octrees remains significant.
In cases where extremely high-fidelity models or large environments are used, the memory demands of the octree representation are sometimes quite high causing the simulation to fail.  
Additionally, we have functionality to spawn props into the environment via the python interface, but the octree-based implementation is not integrated to update and enable viewing those props with the sonar.

Implementing the sonar using direct ray tracing addresses each of the challenges outlined above.
Changing the environment or modifying the sonar resolution parameters no longer requires regeneration of the octree cache. 
Due to the fact that ray tracing happens at each tick, even live changes in the environment (such as a spawned prop) can be observed in the sonar returns. 

To test this, we implemented ray tracing on a simple single-beam sonar (also known as an echo sounder). 
The ray tracing implementation showed significant speed improvements over the octrees, both for the initial run that generates the octrees and subsequent runs that only query them. 
These results are shown in Table \ref{tab:raycast_timing}. 
The increased performance is likely due to the memory-access time needed to read in the cached octree. 
We also confirmed that the ray tracing implementation detects spawned props and works in the complex environments where the octree implementation failed.

Based on the success of these experiments, we plan to develop ray casting implementations for all HoloOcean sonar sensors, including sidescan, imaging, and profiling sonars.

\begin{table}
    \centering
    \begin{tabular}{|c|c|c|}
        \hline
        Run Type & \makecell{Mean Time \\per Tick (s)} & \makecell{Total Time \\ per Tick (s)} \\ \hline
        Octree Caching Run & 0.351 & 178.44 \\ \hline
        Octree Querying Run & 0.055 & 28.187 \\ \hline
        Ray Casting Run & 0.012 & 6.058 \\ \hline
    \end{tabular}
    \caption{Speed test of sonar implementations for a single-beam echo-sounder sonar. Speed was compared for the run that generated the octree, subsequent runs that only query the octrees, and a run using the upgraded ray tracing-based sonar implementation. Simulation was done in our Dam environment for 509 ticks at 30 ticks/sec (0.033 sec per tick), for a real-world equivalent of 16.9667 seconds.}
    \label{tab:raycast_timing}
    \vspace{-2em}
\end{table}

\begin{figure}[b]
    \centering
    \includegraphics[width=0.9\linewidth]{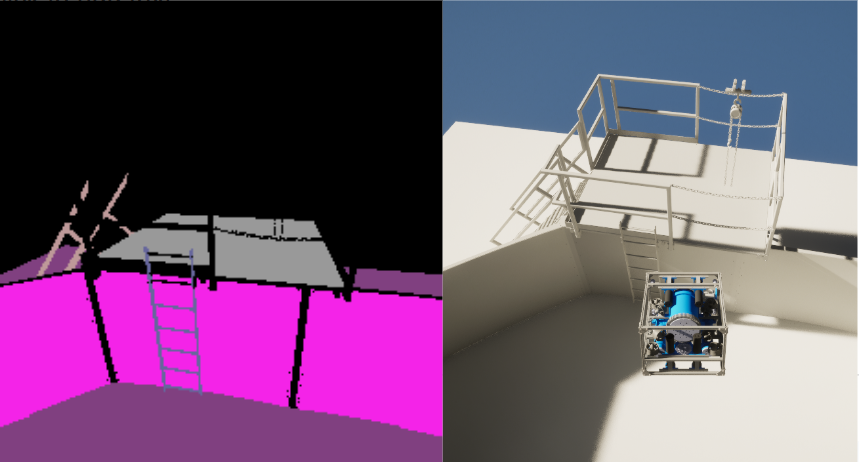}
    \caption{Example output from a semantic camera sensor showing the side of a tank environment.}
    \label{fig:semantic-camera}
\end{figure}

\subsection{Semantic Sensors}
One benefit of simulation is the availability of ground truth information for sensors.
Sensor models in simulation add noise to this ground truth to mimic real world sensor behavior, but the ground truth itself can also be exposed to the user. 
This can be used for training algorithms and is particularly useful for deep learning models.
In visual applications, the desired ground truth is often a pixel-wise segmented image with semantic or instance labels for each region and object in the image.

We are working to add ``semantic sensors" to HoloOcean that automatically return semantically labeled ground truth images based on tags added to objects in the environment. 
These semantic sensors will enable users to train segmentation and detection models on data collected in HoloOcean.
We have implemented semantic sensors for the new camera sensor, depth camera, and LiDAR, and are developing them for the new ray casting sonar sensors. Figure \ref{fig:semantic-camera} illustrates the output from a semantic camera sensor.

\begin{figure}[t]
    \centering 
    \includegraphics[width=1\linewidth]{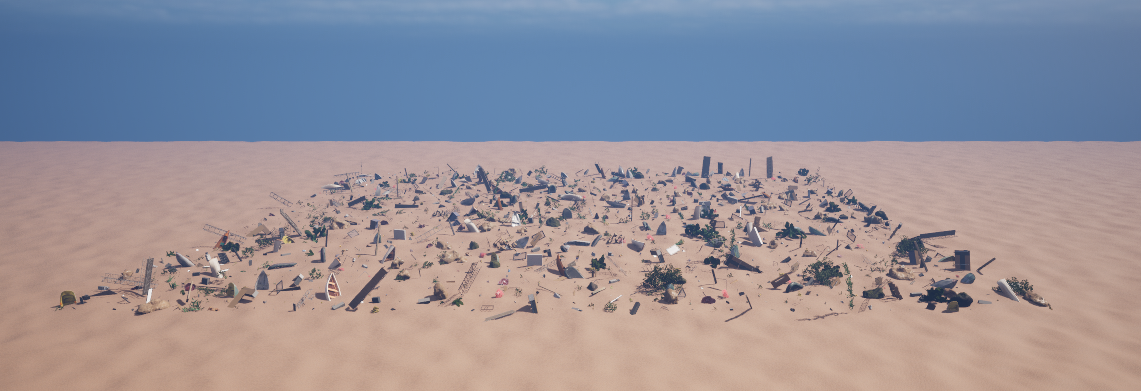}
    \caption{Example environment with randomized object placement made using Python.}
    \label{fig:environment-generation}
\end{figure}
\begin{figure} [t]
    \centering
    \includegraphics[width=1\linewidth]{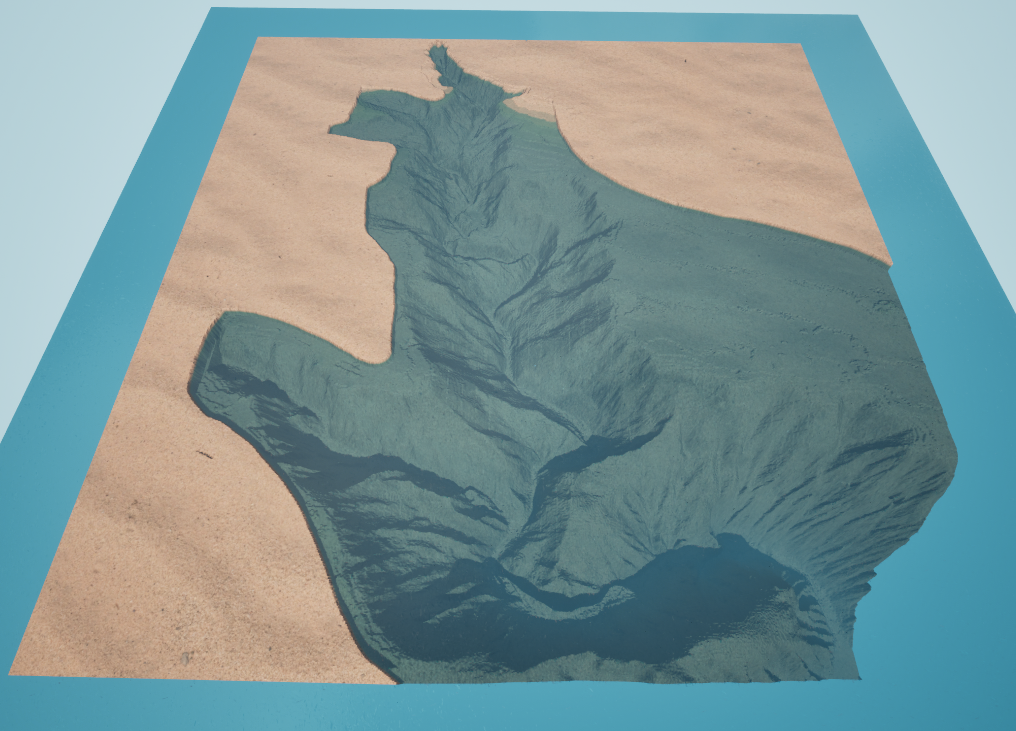}
    \caption{A portion of Monterey Bay loaded into HoloOcean.}
    \label{fig:bathymetry-example}
    \vspace{-2em}
\end{figure}
\subsection{Environment Generation}
HoloOcean currently provides a limited number and variety of environments.
The levels are static and cannot be changed based on user parameters.
Users can create their own levels, but this requires familiarity with UE and can be very time consuming, especially for applications such as model training, where a wide variety of unique environment data is needed. 

We are developing a user-friendly method to add automatically generated environments into HoloOcean. 
These environments will allow the user to specify a variety of parameters, such as terrain type and complexity, as well as presence and density of objects, including natural and man-made structures.

We have successfully utilized Python scripting for UE to randomly load terrain, water, and assets into a world demonstrated in Figure \ref{fig:environment-generation}.
In addition, we are testing a feature that allows users to input bathymetry maps to load real-world locations into HoloOcean. We have loaded a portion of Monterey Bay into HoloOcean as seen in Figure \ref{fig:bathymetry-example}.
These maps are highly detailed and have already been tested to function with our new ray casting sonar methods.

\subsection{Environment Volumetric Effects}
On its first release, HoloOcean's primary objective was to support algorithm development for perception and localization. 
As HoloOcean's user base has grown, additional simulation needs have arisen.
One way we aim to address this gap is by introducing volumetric environmental effects such as ocean currents. We are working on an interface to sample the current for each vehicle from the world based on the vehicle's position at each tick. 

Our implementation of Fossen dynamics for torpedo vehicles includes the ability to specify the direction and magnitude of a current.
However, this parameter is defined on a per-agent basis, and is not consistent across an environment.
We are currently validating a world-level implementation of ocean currents based on a user-defined volumetric object loaded into UE.
The volumetric object specifies a force vector at each location in the world that acts on agents. 
The next steps in this implementation include 1) integrating the volumetric object with the Python client to allow for user configuration when defining the simulation scenario, and 2) sampling the current from the volumetric object at each tick and passing this into the Fossen vehicle dynamics.

Other volumetric effects are being considered for future implementation, including temperature, salinity, and biological and chemical concentrations. 

\subsection{Realistic Waves}
Most marine simulators, including HoloOcean, use a flat plane to represent the surface of the water. They use water shaders to give the appearance of waves and reflections, and usually implement a simple buoyancy system based on the agent's depth relative to the plane.
This approach is not sufficient for many tasks involving surface vessels and underwater vehicles at or near the surface.
Real waves can induce significant motion in the vehicles that affects communication, data collection, and control and localization algorithms.
A flat plane approach cannot capture this complexity in vehicle dynamics. 
Our objective is to implement realistic waves in our HoloOcean environments and to accurately model the surface and subsurface hydrodynamics to enable realistic simulation of surface tasks. 

UE 5 gives us access to several plugins that enable accurate simulation of water surfaces \cite{WaterSystem, WaterlinePro}.
These plugins use Gerstner or FFT waves to produce highly realistic water surfaces for both open water and coastline interactions. They also incorporate advanced shaders for visual effects such as simulating reflection, caustics, and interactions with vehicles.

We are currently testing two water plugins for UE 5 to determine if the waves they produce will work with our sensor and vehicle implementations.
Initial testing with UE's native Water System plugin disrupts the octrees of the sonar implementation, but does not interfere with the new direct ray casting sonar implementation.
Next steps include improving our buoyancy system by sampling the height of the wave surface at the agent's location and accounting for agent geometry. 
We are also looking into modeling other wave-associated hydrodynamic forces, such as those induced on a vehicle just under the surface by the back-and-forth movement of the water. 

\section{Conclusion}
\label{sec:conclusion}

HoloOcean 2.0 continues HoloOcean's success by adding improved visual rendering with Unreal Engine 5.3, high-fidelity vehicle dynamics using Fossen's models, and support for ROS2. Features that are nearly complete include new sensors such as depth camera and LiDAR, sonar implementations using ray casting, and semantic ground truth data for sensors. Other features under investigation include environment generation, water currents, and waves. 

Examples and documentation can be found at our docs page \url{https://byu-holoocean.github.io/holoocean-docs/} which will be updated when the features in progress are complete.

The demands of marine robotics simulators will continue to evolve as the field of robotics advances. We hope to continue development of HoloOcean with updated features as a support to the marine robotics research community as the field progresses.

\bibliographystyle{style/IEEEtranN} %
{\footnotesize
\bibliography{references/IEEEabrv,references/strings-full,references/library}}

\end{document}